\documentclass[10pt,letterpaper]{article}
\usepackage[top=0.85in,left=2.75in,footskip=0.75in]{geometry}

\usepackage{amsmath,amssymb}

\usepackage{changepage}

\usepackage{textcomp,marvosym}

\usepackage{cite}

\usepackage{nameref,hyperref}

\usepackage[right]{lineno}

\usepackage{microtype}
\DisableLigatures[f]{encoding = *, family = * }

\usepackage[table]{xcolor}

\usepackage{array}

\newcolumntype{+}{!{\vrule width 2pt}}

\newlength\savedwidth

\raggedright
\usepackage[aboveskip=1pt,labelfont=bf,labelsep=period,justification=raggedright,singlelinecheck=off]{caption}

\makeatletter
\renewcommand{\@biblabel}[1]{\quad#1.}
\makeatother

\usepackage{lastpage,fancyhdr,graphicx}
\usepackage{epstopdf}
\fancyheadoffset[L]{2.25in}
\fancyfootoffset[L]{2.25in}
\usepackage{tikz}

\begin{document}
\vspace*{0.2in}

\begin{flushleft}
{\Large
\textbf\newline{Correcting for heterogeneity in real-time epidemiological indicators} 
}
\newline
\\
Aaron Rumack\textsuperscript{1*},
Roni Rosenfeld\textsuperscript{1},
F. William Townes\textsuperscript{2}
\\
\bigskip
\textbf{1} Machine Learning Department, Carnegie Mellon University, Pittsburgh,
Pennsylvania, United States of America \\
\textbf{2} Department of Statistics \& Data Science, Carnegie Mellon University,
Pittsburgh, Pennsylvania, United States of America \\
\bigskip

*arumack@andrew.cmu.edu

\end{flushleft}

\section*{Abstract}
Auxiliary data sources have become increasingly important in epidemiological surveillance, as they are often available at a finer spatial and temporal resolution, larger coverage, and lower latency than traditional surveillance signals. We describe the problem of spatial and temporal heterogeneity in these signals derived from these data sources, where spatial and/or temporal biases are present. We present a method to use a ``guiding'' signal to correct for these biases and produce a more reliable signal that can be used for modeling and forecasting. The method assumes that the heterogeneity can be approximated by a low-rank matrix and that the temporal heterogeneity is smooth over time. We also present a hyperparameter selection algorithm to choose the parameters representing the matrix rank and degree of temporal smoothness of the corrections. In the absence of ground truth, we use maps and plots to argue that this method does indeed reduce heterogeneity. Reducing heterogeneity from auxiliary data sources greatly increases their utility in modeling and forecasting epidemics.



\section{Introduction}

Understanding the burden of epidemics is a critical task for both public health officials and modelers. However, traditional surveillance signals are often not available in real-time, due to delays in data collection as well as data revisions. Alternative data sources can provide more timely information about an epidemic's current state, which can be useful for modeling and forecasting. We can use these data sources to create \textit{indicators}, which provide a single number quantifying some measure of epidemic burden for a given location and time. An indicator usually estimates the disease burden at a certain severity level (e.g. symptomatic infections, hospitalizations) when the ground truth is unobserved. During the COVID-19 pandemic, the Delphi group published a repository of several real-time indicators of COVID-19 activity \cite{reinhart2021}.

Many, if not all, of these indicators suffer from heterogeneity. That is, the relationship between the indicator and unobserved ground truth changes over space or time. To define heterogeneity, let $X \in \mathbb{R}^{N \times T}$ be the matrix containing the indicator values for $N$ locations and $T$ time values, and $Z \in \mathbb{R}^{N \times T}$ be the matrix containing the corresponding ground truth values. We say that spatial heterogeneity is present when $$\mathbb{E}[X_{i_1t}] - Z_{i_1t} \neq \mathbb{E}[X_{i_2t}] - Z_{i_2t} \text{ for some } i_1 \neq i_2, t.$$
Likewise, temporal heterogeneity is present when $$\mathbb{E}[X_{it_1}] - Z_{it_1} \neq \mathbb{E}[X_{it_2}] - Z_{it_2} \text{ for some } i,t_1 \neq t_2.$$

Note that we define heterogeneity not simply as a bias in the indicator, but rather that the bias is dependent on location or time. The causes of heterogeneity vary depending on the indicator, but we can consider as an example an indicator based on insurance claims that seeks to estimate incidence of COVID-19 outpatient visits. Insurance claims could be higher relative to COVID-19 incidence in locations where the population in the insurance dataset is older, or where the doctors have more liberal coding policies in labeling a probable COVID case. Even the signal of reported cases, which purportedly reflects COVID-19 infections directly, will suffer from heterogeneity. If a few locations suffer from a shortage of tests, or from a new strain which tests are less accurate in detecting or that has a different fraction of symptomatic cases, those locations will have a different relationship between reported cases and true cases. Similar causes can result in temporal heterogeneity. Test shortages, changing demographics, coding practices can also vary over time within a single location. For example, spatial heterogeneity has been documented in CDC's ILINet due to different mixtures of reporting healthcare provider types in the network \cite{delphi_ilinearby}.

We use real-time indicators for three main functions: modeling the past, mapping the present, and forecasting the future. Correcting for heterogeneity is important for all of these applications. Any statistical conclusions we make about spatiotemporal spread of a disease may be distorted if the underlying data is subject to heterogeneity. In the presence of spatial heterogeneity, the indicator values are not comparable across locations, and a choropleth map displaying the current values of the indicator will be misleading. Similarly, in the presence of temporal heterogeneity, displaying a time series of the indicator may be misleading. Heterogeneity affects forecasts as well, as biases in the features of a forecasting model will lead to forecast inaccuracy. Our goal is to remove heterogeneity in an indicator in order to make it more reliable for these three uses.

Heterogeneity has been described and modeled in the field of econometrics \cite{tsionas2019}. Nearly all of the work involving heterogeneity in econometrics deals with the implications in regression. If only spatial heterogeneity is present, then a fixed or random effects model can be used \cite{greene2005,wang2010}. Others have developed parametric methods that assume heterogeneity is also time-varying \cite{kutlu2019}. The main reason that these methods cannot be transferred to our domain is that they identify heterogeneity only through strict assumptions on the error terms in the regression model. Additionally, we are not performing regression in our application. Rather, we are trying to remove the heterogeneity in the indicator.

A challenge of correcting for heterogeneity is that the problem doesn't have a clear formulation. In nearly every practical application, we lack access to the ground truth and our best option is to compare our indicator with another signal that is a noisy estimate of the ground truth, and often suffers from heterogeneity itself. We will call this signal a ``guide'' to emphasize that it is not a target for prediction. We believe that the indicator is strongly related with the guide, so they should be correlated across time and space. However, they don't measure the same value, so the correlation should not be $1$ even in the absence of noise. Another challenge is that we present the problem in a retrospective setting, without a clear division for training and testing.

In this paper, we investigate removing heterogeneity from two indicators using a different guide for each. The first indicator is based on insurance claims data, and we use reported cases as a guide signal. The second indicator is based on Google search trends of specific queries related to COVID-19. We use the COVID-19 Trends and Impact Surveys (CTIS) as a guide. All of these signals (indicators and guides) are available in COVIDCast \cite{reinhart2021}.

Because heterogeneity is present in a wide variety of indicators, we desire a solution that is general and flexible. Another desired property is that the temporal corrections are smooth across time, because we want to accommodate situations where the relationship between the indicator and guide can drift slowly over time. The model should be flexible enough to allow for abrupt changes, but these should be limited in number. If the corrections are jagged in time, the model may be overadjusting to the guide signal rather than identifying and removing the true heterogeneity.

Lastly, the method should generalize well to a variety of indicators and guides. It should not rely on specific domain knowledge of a single indicator-guide pair because we want the method to be applicable to any current or future indicator and guide. If we believe the indicator and guide have a stronger relationship, then we might want the model to use the guide matrix more and make a stronger bias correction. If we believe that there is more noise in the guide variable, that heterogeneity is mild, or that the inherent signals are more divergent, we might want the model to make a weaker bias correction. Additionally, the temporal smoothing constraint will be stronger or weaker, depending on the application.

The model should have hyperparameters to control the strength of the guide signal in fitting as well as the strength of the temporal smoothness constraint. These can be conceptualized as ``knobs''. For the indicator-guide relationship, the knob turns between one extreme of not using the guide signal at all and the other extreme of fitting maximally to the guide signal (in some models, fitting exactly to the guide signal). For the temporal smoothness constraint, the knob turns between the extremes of applying no smoothing and enforcing a constant temporal correction factor across time.

In the rest of this paper, we will provide three methods to correct for heterogeneity for a general indicator and guide signal. We then demonstrate their performance in simulated experiments and on several actual epidemiological data sources.

\section{Methods}
Let $X \in \mathbb{R}^{N \times T}$ be the matrix containing the indicator values for $N$ locations and $T$ time points, and $Y \in \mathbb{R}^{N \times T}$ be the matrix containing the corresponding guide values. We want to transform $X$ to a matrix $\tilde{X}$, with the spatial and temporal biases mitigated. As mentioned above, the simplest way to do so is to set $\tilde{X} = Y$, but this is the most extreme version of overadjustment and removes any unique information contained in $X$. We will present three methods to remove heterogeneity by using $Y$ as a guide. The first uses a simple low-rank approximation, and the second and third add elements which ensure that the biases removed are smooth in time. In all of our methods, we detect heterogeneity by examining the difference $Y-X$. We assume that the signal in this difference matrix is the heterogeneity between $X$ and $Y$.

\subsection{Bounded Rank Approach}
In this approach, we assume that the heterogeneity between $X$ and $Y$ is of low rank. We begin without making any assumptions on the smoothness of the temporal biases. Therefore, we solve the following optimization:
$$
\hat{A}, \hat{B} = \arg \min_{A,B} \|(X + AB^T) - Y \|_F^2,
$$
where $A \in \mathbb{R}^{N \times K}$, $B \in \mathbb{R}^{T \times K}$, $K \leq \text{min}(N,T)$, and $\|\cdot\|_F$ is the Frobenius norm. This optimization can be solved by performing singular value decomposition on the difference matrix $Y - X$ and keeping the vectors with the $K$ highest singular values. The corrected matrix is $\tilde{X} = X + AB^T$.

\subsection{Fused Lasso Approach}
In addition to the low rank assumption, here we further assume that the temporal biases are mostly piecewise constant over time. Therefore, we solve the following optimization:
$$
\hat{A}, \hat{B} = \arg \min_{A,B} \|(X + AB^T) - Y \|_F^2 + \lambda \|\Delta_t B\|_1,
$$
where $A \in \mathbb{R}^{N \times K}$, $B \in \mathbb{R}^{T \times K}$, and $K \leq \text{min}(N,T)$, and $\Delta_t B$ contains the first differences of B along the time axis. The $\Delta_t B$ penalty is inspired by the fused lasso \cite{tibshirani2005} and encourages $B$ to be piecewise constant along the time axis.

We solve this optimization using penalized matrix decomposition algorithms described in \cite{witten2009}. We reproduce the algorithm as applicable to our case here:
\begin{enumerate}
    \item Let $Z^{1} = Y - X$.
    \item For $k = 1, \dots, K$:
    \begin{enumerate}
        \item Initialize $v_k$ to have $L_2$ norm $1$.
        \item Iterate until convergence:
        \begin{enumerate}
            \item If $v_k = 0$, then $u_k = 0$. Otherwise, let $u_k = \frac{Z^kv_k}{\|Z^kv_k\|_2}$.
            \item Let $v_k$ be the solution to
            $$\min_v \frac{1}{2}\|Z^{kT}u_k - v\|_2^2 + \lambda \sum_{j=2}^T \|v_j - v_{j-1}\|_1.$$
        \end{enumerate}
        \item Let $d_k = u_k^TZ^kv_k$.
        \item Let $Z^{k+1} = Z^k - d_ku_kv_k^T$.
    \end{enumerate}
    \item $A$ is the matrix whose $k^{th}$ column is $d_ku_k$, and $B$ is the matrix whose $k^{th}$ column is $v_k$.
\end{enumerate}

Step 2b) ii) is a fused lasso problem and can be solved using the alternating direction method of multipliers (ADMM) \cite{boyd2011}. All of the other steps are trivial to compute.

This optimization has two hyperparameters which can be considered as ``knobs'': $K$ and $\lambda$. The matrix rank $K$ controls the degree to which we match the guiding signal $Y$. When $K = 0$, we keep $X$ exactly as is and apply no correction. As $K$ increases, we use more information from $Y$, and when $K = \min(N,T)$, we transform $X$ to equal $Y$ exactly (when $\lambda = 0$). The lasso penalty $\lambda$ enforces smoothness along the time axis of $B$. At $\lambda = 0$, we apply no smoothing at all, and the model is equivalent to the Bounded Rank Model above. As $\lambda$ approaches $\infty$, $B$ contains a constant value across each row.

\subsection{Basis Spline Approach}
An alternative way to enforce smoothness on the temporal bias correction is to transform the temporal corrections by using B-spline basis functions. These functions $S$ are determined by setting the polynomial degree $d$ and a set of knots $\{t_1,\dots, t_m\}$ \cite{deboor2001}:
$$S_{i,0}(x) = 1, \text{ if } t_i \leq x < t_{i+1}, \text{ otherwise } 0,$$
$$S_{i,k}(x) = \frac{x - t_i}{t_{i+k}-t_i}S_{i,k-1}(x) + \frac{t_{i+k+1} - x}{t_{i+k+1} - t_{i+1}}S_{i+1,k-1}(x),$$
for $i \in \{1,\dots,m\}$ and $k \in \{1,\dots d\}$.
We can use these basis functions to create a fixed spline transformation matrix $C \in \mathbb{R}^{L \times T}$, where $C_{i,t} \equiv S_{i,d}(t)$ and $L$ is a function of $d$ and $m$.

We now solve the following optimization:
$$
\hat{A}, \hat{B} = \arg \min_{A,B} \|(X + AB^TC) - Y \|_F^2,
$$
where $A \in \mathbb{R}^{N \times K}$, $B \in \mathbb{R}^{L \times K}$, and $K \leq \text{min}(N,L)$, and $C$ is the spline transformation matrix determined by the given polynomial degree and knots. This problem can be reformulated and solved by reduced rank regression, using the algorithm described in \cite{mukherjee2011}. In this approach, we do not need to apply a penalty to the components of $B$; the spline basis transformation will ensure that the temporal correction matrix $B^TC$ is smooth.

In this approach, the hyperparameter $K$ is understood the same way as above. The temporal smoothing hyperparameters are different, however. The degree of smoothing is determined by the polynomial degree $d$ and knots $t$. For simplicity, we will set $d$ as a constant 3; this results in the commonly used cubic spline transformation. We will also enforce that the knots are uniformly spaced, leaving us with the knot interval as the only temporal hyperparameter. The larger the knot interval, the smoother the temporal corrections will be. Note that due to the transformation matrix $C$, we are no longer able to fit $\tilde{X} = Y$ exactly, even with unbounded $K$.

We note that we can parameterize the Basis Spline Approach to be equivalent to the Fused Lasso Approach. By setting the basis spline degree to be $d=0$, the spline transformation matrix $C$ results in a vector that is piecewise constant. If we place a knot at every time point and apply an $\ell_1$ penalty to the first differences of the spline components, then the Basis Spline Approach is equivalent to the Fused Lasso Approach. Analogous equivalences hold for higher order splines. If the basis spline degree is $d=1$, the method is equivalent to trend filtering \cite{kim2009}, and so on for higher polynomial degrees.

\subsection{Preprocessing Indicator Values}
\label{ssec:preprocess}
All of the models above assume that the heterogeneity corrections should be additive, that is, $\tilde{X} = X + AB^T$. Depending on the application, it may be more reasonable to apply a multiplicative correction. In such a case, we can fit the models using $\log X$ and $\log Y$. If $X$ or $Y$ contain zeros, then we can add a pseudocount and fit using $\log (X+\epsilon)$. We optimize
$$
\min_{A,B} \| (\log X + AB^T) - \log Y\|_2^F
$$
for the Bounded Rank Model, and the temporal penalties are straightforward for the Fused Lasso and Basis Spline models. Our corrected indicator is $\tilde{X} = X \odot \exp (AB^T)$, where $\odot$ represents the Hadamard product and exponentiation is element-wise. One caveat to note is that the optimization minimizes the mean squared error between the indicator and guide on the log scale.

\subsection{Hyperparameter Selection}
\label{ssec:hyperparameter}
Each of our three models has one or two hyperparameters that control how the guide signal is used. A user may have domain knowledge which suggests that a certain rank is appropriate, in which case, $K$ can be selected manually. A rank could also be selected via various heuristics, such as an elbow plot of the principal components of $Y-X$. Alternatively, multiple values of $K$ could be selected for sensitivity analysis. In this section, we provide a quantitative method of selecting hyperparameters as a default option, as an alternative to manual selection.

In our setting, several factors complicate the usually straightforward application of cross validation. First, the data is structured in a two dimensional matrix. Our optimization method does not allow missingness in the matrices, so we cannot simply remove a random subset of data and run the optimization procedure. We can remove entire columns (time points) either randomly or in blocks, but we will need to interpolate the values for the missing time points. We can use mean squared error between $\tilde{X}$ and $Y$ as the error metric, but it is not clear that this is an ideal choice. The indicator and guide measure different quantities and we do not believe or wish that success is defined as matching $\tilde{X}$ and $Y$.

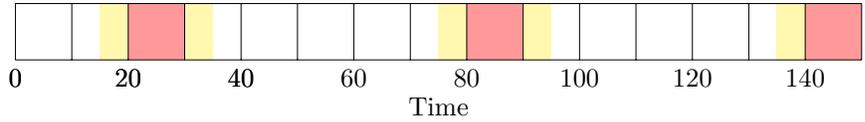
\begin{figure}
\centering
\begin{tikzpicture}[scale=0.75]
\fill[red!40!white] (2,0) rectangle (3,1);
\fill[red!40!white] (8,0) rectangle (9,1);
\fill[red!40!white] (14,0) rectangle (15,1);
\fill[yellow!40!white] (1.5,0) rectangle (2,1);
\fill[yellow!40!white] (3,0) rectangle (3.5,1);
\fill[yellow!40!white] (7.5,0) rectangle (8,1);
\fill[yellow!40!white] (9,0) rectangle (9.5,1);
\fill[yellow!40!white] (13.5,0) rectangle (14,1);
\draw[step=1cm,black,very thin] (0,0) grid (15,1);
\filldraw[black] (0,0) circle (0pt) node[anchor=north]{0};
\filldraw[black] (2,0) circle (0pt) node[anchor=north]{20};
\filldraw[black] (4,0) circle (0pt) node[anchor=north]{40};
\filldraw[black] (6,0) circle (0pt) node[anchor=north]{60};
\filldraw[black] (8,0) circle (0pt) node[anchor=north]{80};
\filldraw[black] (10,0) circle (0pt) node[anchor=north]{100};
\filldraw[black] (12,0) circle (0pt) node[anchor=north]{120};
\filldraw[black] (14,0) circle (0pt) node[anchor=north]{140};
\filldraw[black] (0,0) circle (0pt) node[anchor=north]{0};
\filldraw[black] (2,0) circle (0pt) node[anchor=north]{20};
\filldraw[black] (4,0) circle (0pt) node[anchor=north]{40};
\filldraw[black] (7.5,-0.5) circle (0pt) node[anchor=north]{Time};
\end{tikzpicture}

\caption{\small We use cross-validation for hyperparameter selection. The red blocks (ten days each) are held out for testing, and the yellow blocks (five days each) are held out to reduce dependencies between the training data and the test data. We repeat for 6 folds.}
\label{fig:cv_diagram}
\end{figure}

\begin{figure}
    \centering
    \includegraphics[width=0.9\linewidth]{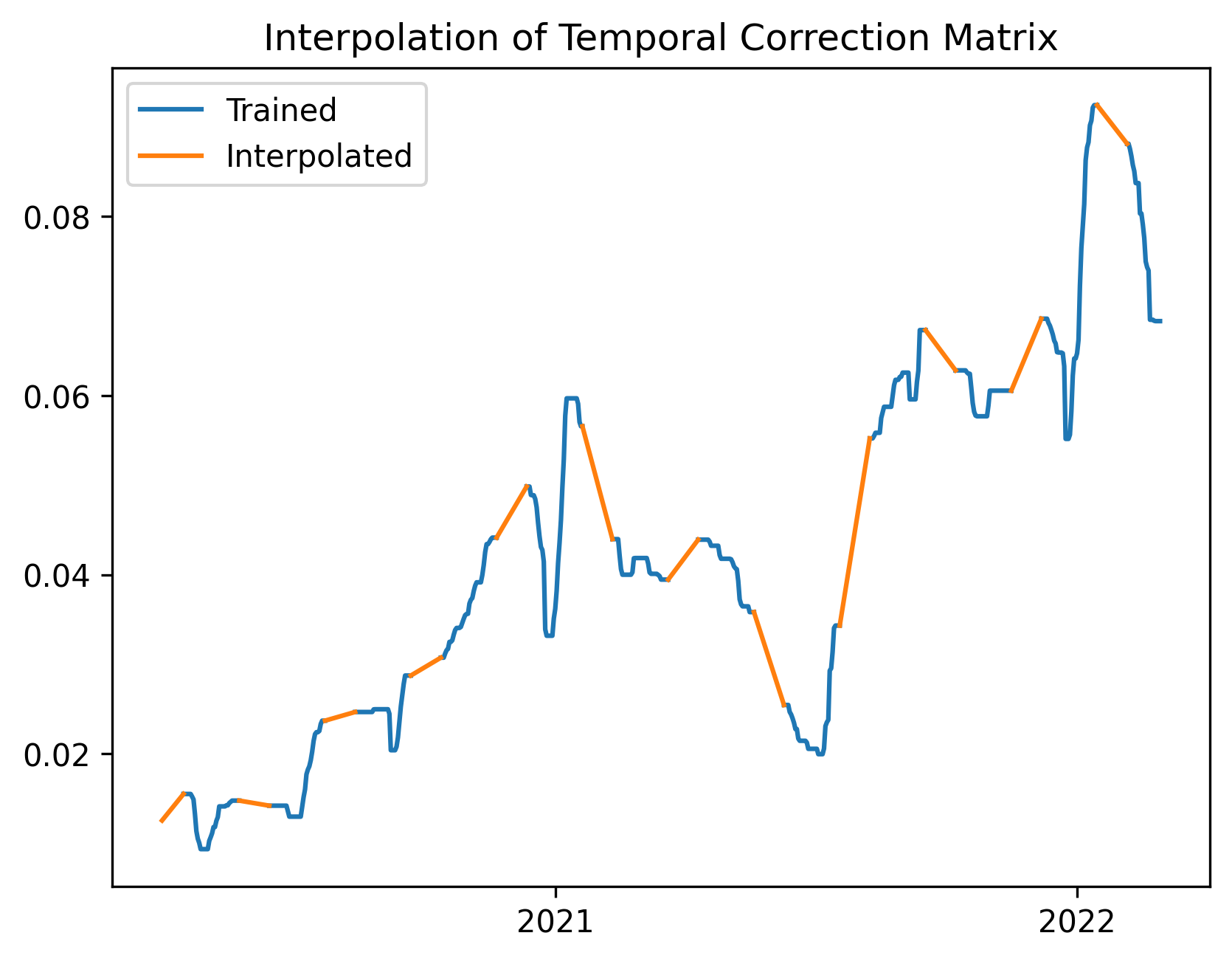}
    \caption{\small We need to interpolate the test indices of the temporal adjustment matrix $B$ in order to calculate $\tilde{X}$. We do this by linear interpolation between the values of $B$ on the boundaries of the blocks of training indices. This figure shows interpolation for a single column of matrix $B$, as an example.}
    \label{fig:interpolate_b}
\end{figure}

Despite these challenges, we will select hyperparameters by using a cross validation framework with mean squared error as the error metric. In order to reduce the temporal dependencies inherent in the data, we leave out blocks of time for testing, as illustrated in Fig~\ref{fig:cv_diagram}. We use linear interpolation to populate the rows of $B$ in the test set, as illustrated in Fig~\ref{fig:interpolate_b}. Our error metric is the mean squared error between $\tilde{X}$ and $Y$ on the test set.

In the penalized regression context, it is common to apply the ``one standard error rule'' to cross validation, in which we select the most parsimonious model whose cross validation error is within one standard error of the the minimum cross validation error across all models \cite{hastie2009}. A common justification for this rule is that the cross validation errors are calculated with variance, and it is preferable to take a conservative approach against more complex models \cite{hastie2009}. Our setup provides further motivation to apply this rule. Unlike in standard cross validation, our goal is not to find the model which fits best to $Y$, but rather to use $Y$ as a guiding signal to mitigate heterogeneity. Additionally, there is likely a slight dependence between the training data and test data due to the temporal structure of the data. Applying the ``one standard error rule'' will prevent overadjustment to $Y$.

In order to use the ``one standard error rule'', we will need to calculate the number of parameters for a given model. For the Bounded Rank and Basis Spline models, this is straightforward. For the Bounded Rank Model, the number of degrees of freedom is $K(N+T-1)$, and for the Basis Spline Model, it is $K(N+L-1)$, where $L$ is the dimensionality of the basis spline transformation matrix $C$. For the Fused Lasso Model, we cannot simply calculate the number of entries in the matrices $A$ and $B$. We will use a result that applies to generalized lasso problems under weak assumptions \cite{tibshirani2012}. In our case, we will estimate the degrees of freedom in matrix $B$ as $\|\Delta_t B\|_0$, or the count of non-zero successive differences along the time axis of $B$. The total degrees of freedom for the Fused Lasso Model is $K(N-1) + \|\Delta_t B\|_0$.

We note that the theorem in \cite{tibshirani2012} applies only to generalized lasso problems, and in our case, we use an iterative approach of which the fused lasso is just a subroutine. Therefore, the results may not hold precisely in our case. However, we are using the ``one standard error rule'' merely as a heuristic, and we do not require absolute accuracy in estimating the degrees of freedom.

We reiterate that this is a general rule, and the user can use any rule to select hyperparameters. If a user has domain knowledge which suggests that a certain rank is appropriate, then they could simply select that rank. If a user wants a more parsimonious model, they could use a two standard error rule or a three standard error rule. In some cases, the cross validation error may have a clear elbow, which could suggest an ideal rank. The ``one standard error rule'' is used simply as a baseline when no obvious choice exists.

\section{Results}

\subsection{Simulation Experiments}
We first performed experiments on simulated data, where the true rank of the difference matrix $Y-X$ was known. We fit each of the three models to the difference matrix and evaluate performance through cross validation. The simulation setup is as follows:
\begin{enumerate}
    \item Generate $A$ as a $N \times K$ matrix, where $A_{ij} \overset{\text{iid}}{\sim} \text{Unif}(-1,1)$. 
    \item Generate $B$ as a $T \times K$ matrix. For each column $k$, select nine random breakpoints $(b^k_1,...,b^k_9)$ between $1$ and $T-1$. Set $b^k_0 = 0$ and $b^k_10 = T$. Set $B_{b^k_i:b^k_{i+1},k}$ to be a random constant between $0$ and $1$. Thus each column of $B$ is piecewise constant with $10$ pieces.
    \item Let $C = AB^T$, then normalize to have standard deviation $1$ across all elements.
    \item Let the simulated difference matrix $Y-X$ be $D$, where $D_{ij} = C_{ij} + \epsilon_{ij}$ and $\epsilon_{ij} \overset{\text{iid}}{\sim} N(0,\sigma=0.1)$. Note that we do not have to simulate $X$ or $Y$ individually, only the difference.
\end{enumerate}
In our simulations, we used $N=51$ and $T=699$ as in our real-world analysis. We experimented with $K \in \{5, 10, 40\}$. For the simulations, we will discuss the Fused Lasso Model (including $\lambda=0$, which is equivalent to the Bounded Rank Model). The Basis Spline Model does not perform well in the simulations, as will be discussed in the next section.

\begin{figure}
    \centering
    \includegraphics[width=0.7\linewidth]{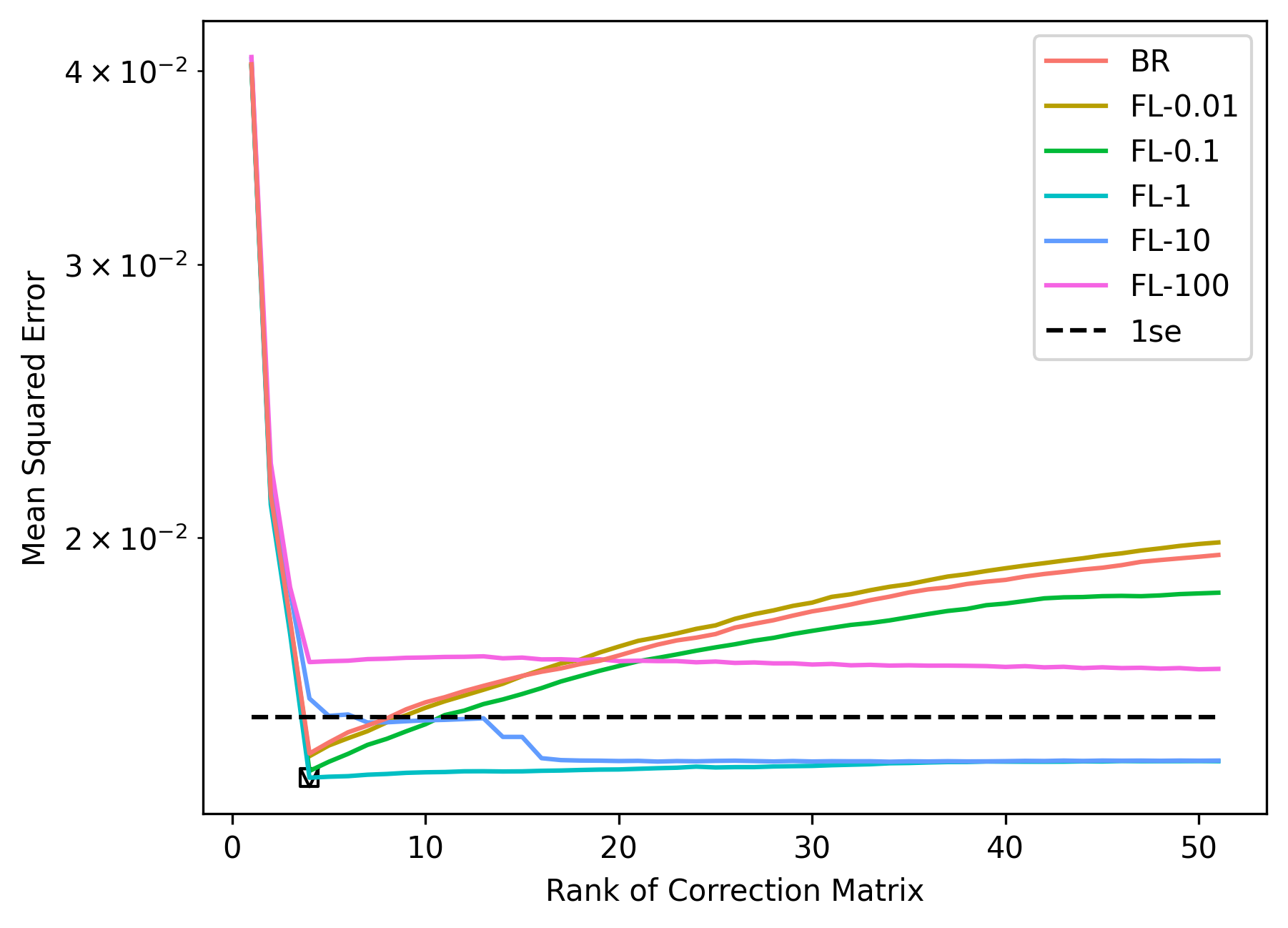}
    \caption{\small Although the true rank of the correction matrix is $5$, the optimal rank selected is $4$ and $\lambda = 1$. For the Bounded Rank Model (BR) and small values of $\lambda$ for the Fused Lasso Model (FL), a clear overfitting curve appears. Even when applying the one standard error rule (1se), the same model is selected (as denoted by the square and triangle).}
    \label{fig:cv_sim5}
\end{figure}

\begin{figure}
    \centering
    \includegraphics[width=0.7\linewidth]{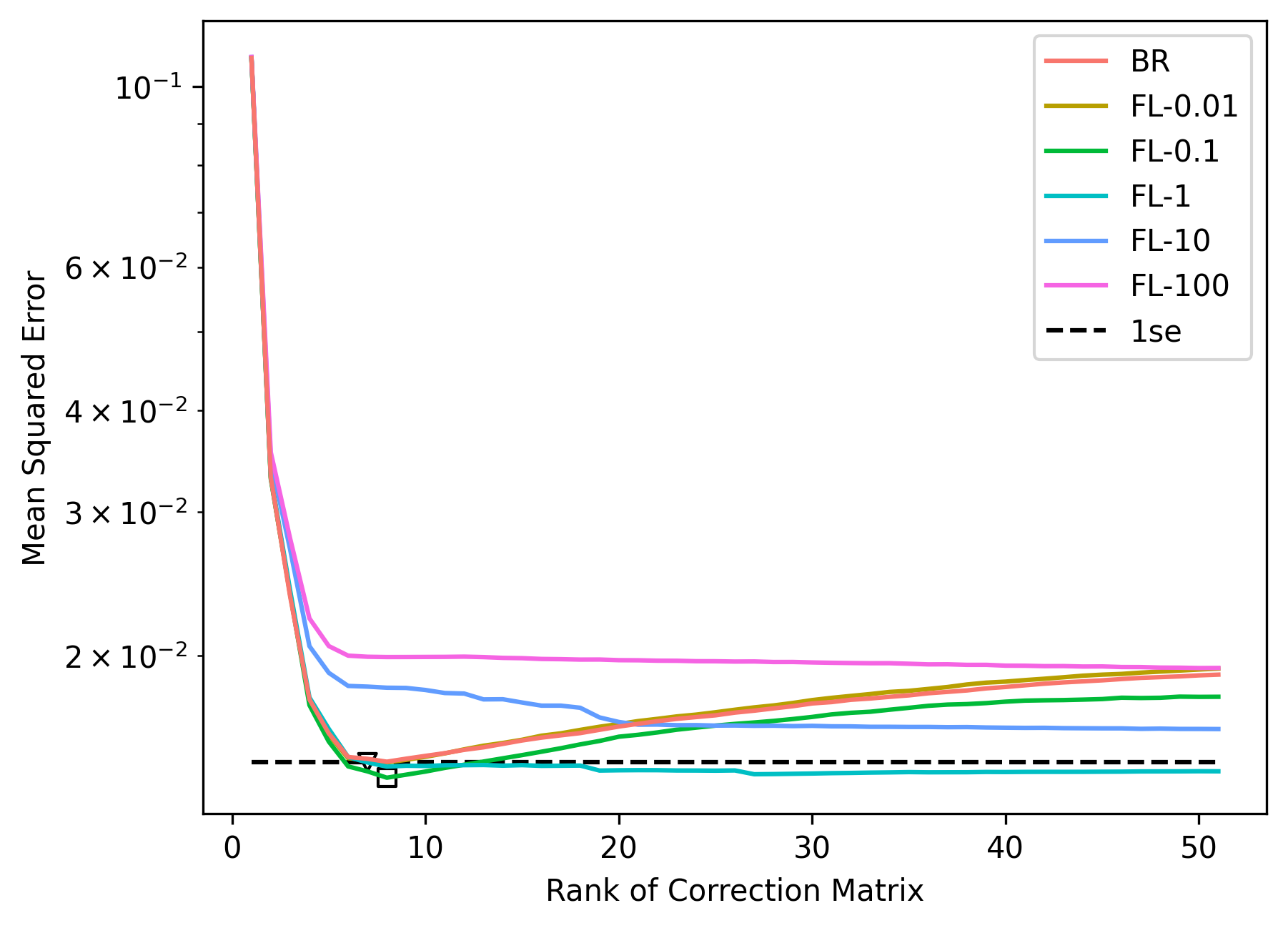}
    \caption{\small Although the true rank of the correction matrix is $10$, the optimal rank selected is $8$ and $\lambda = 0.1$ (denoted by the square). For the Bounded Rank Model (BR) and small values of $\lambda$ for the Fused Lasso Model (FL), a clear overfitting curve appears. When applying the one standard error rule (1se), the model selected has rank $7$ and $\lambda=1$ (denoted by the triangle).}
    \label{fig:cv_sim10}
\end{figure}

For $K=5$, the rank selected by cross validation is $4$, as shown in Fig~\ref{fig:cv_sim5}. In applications where the signal-to-noise ratio is low, our methods will have difficulty in detecting all of the heterogeneity. In this simulation, we can decompose the signal and noise exactly, and perform SVD on the signal and noise separately to determine the signal-to-noise ratio. The first $4$ singular values of the signal matrix $C$ are larger than those of the noise matrix $\epsilon$, but the remainder are smaller. We do not see a strong signal in all $5$ singular values partially because the rows of $AB^T$ were not constructed to be orthogonal. This supports the hypothesis that the optimal rank was not found due to the low signal-to-noise ratio.  We see a similar pattern for $K=10$ (Fig~\ref{fig:cv_sim10}), where the rank selected by cross validation is $8$ and the first $8$ singular values of $C$ are larger than those of $\epsilon$. Once $K$ exceeds the optimal rank, the cross validation error slowly increases, just as would be expected if the model were overfitting.

\begin{figure}
    \centering
    \includegraphics[width=0.7\linewidth]{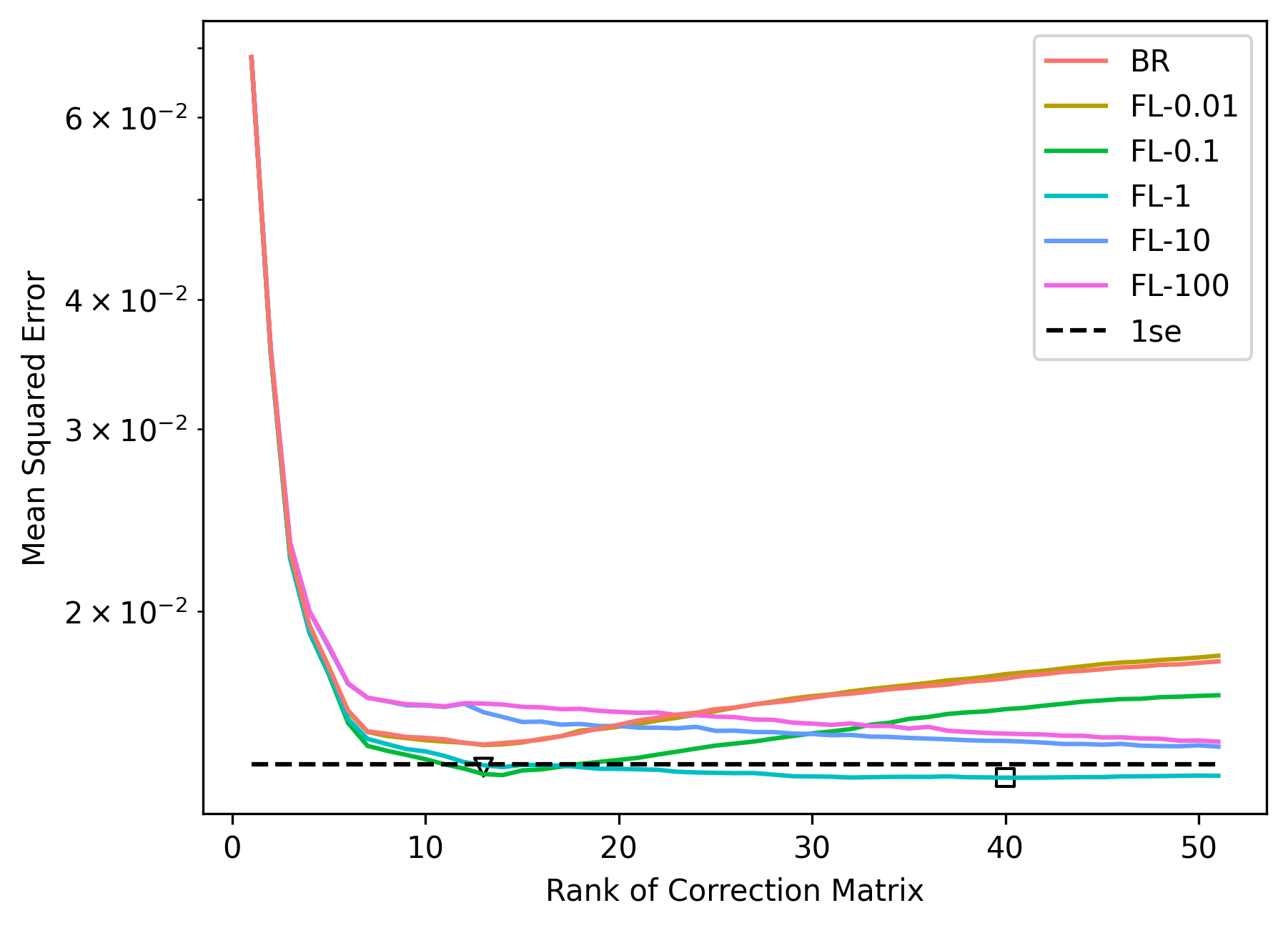}
    \caption{\small The true rank of the correction matrix is $40$, and the optimal rank selected is $40$ with $\lambda = 1$ (denoted by the square). For the Bounded Rank Model (BR) and small values of $\lambda$ for the Fused Lasso Model (FL), a clear overfitting curve appears with the minimum significantly lower than the optimal rank. When applying the one standard error rule (1se), the rank selected is $13$ with $\lambda = 1$ (denoted by the triangle).}
    \label{fig:cv_sim40}
\end{figure}

Fig~\ref{fig:cv_sim40} shows that for $K=40$, the rank with the minimum cross validation error is indeed $40$, using the Fused Lasso Model with $\lambda=1$. Even though the signal-to-noise ratio is higher than 1 for only the first $11$ singular values, the correct rank is selected. This is a somewhat surprising result, and might be attributable to the penalty encouraging the rows of $B$ to be piecewise constant. This may allow the model to detect even parts of the signal that are weaker than the noise.

\subsection{COVID-19 Insurance Claims and Reported Cases}
Insurance claims are a useful data source in modeling and forecasting epidemics. They provide information about how many people are sick enough to seek medical care, which is potentially more useful than simply the number of people who are infected but potentially asymptomatic. They can also be available at high geographic and temporal resolution, as well as cover a large proportion of the total population. In this section, we will use a dataset of aggregated insurance claims provided by Optum. The signal is the fraction of all outpatient claims with a confirmed COVID-19 diagnosis code, followed by smoothing and removal of day-of-week effects \cite{delphi_epidata_api_2020}. Despite the advantages of claims datasets, they are often subject to spatial and temporal heterogeneity, as we will demonstrate.

We used reported COVID-19 cases from Johns Hopkins \cite{dong2020} as our guide signal to correct for heterogeneity in the insurance claims signal. As in the simulation experiments, we used the hyperparameter selection scheme described in Section~\ref{ssec:hyperparameter}. Because we believe that the effects of heterogeneity here are multiplicative rather than additive, we applied preprocessing steps as described in Section~\ref{ssec:preprocess}. We set $X$ to be the log of the insurance claims signal, and $Y$ to be the log of the reported cases signal, each with a pseudocount of $\epsilon=1$ to account for zeros.

\begin{figure}
    \centering
    \includegraphics[width=0.7\textwidth]{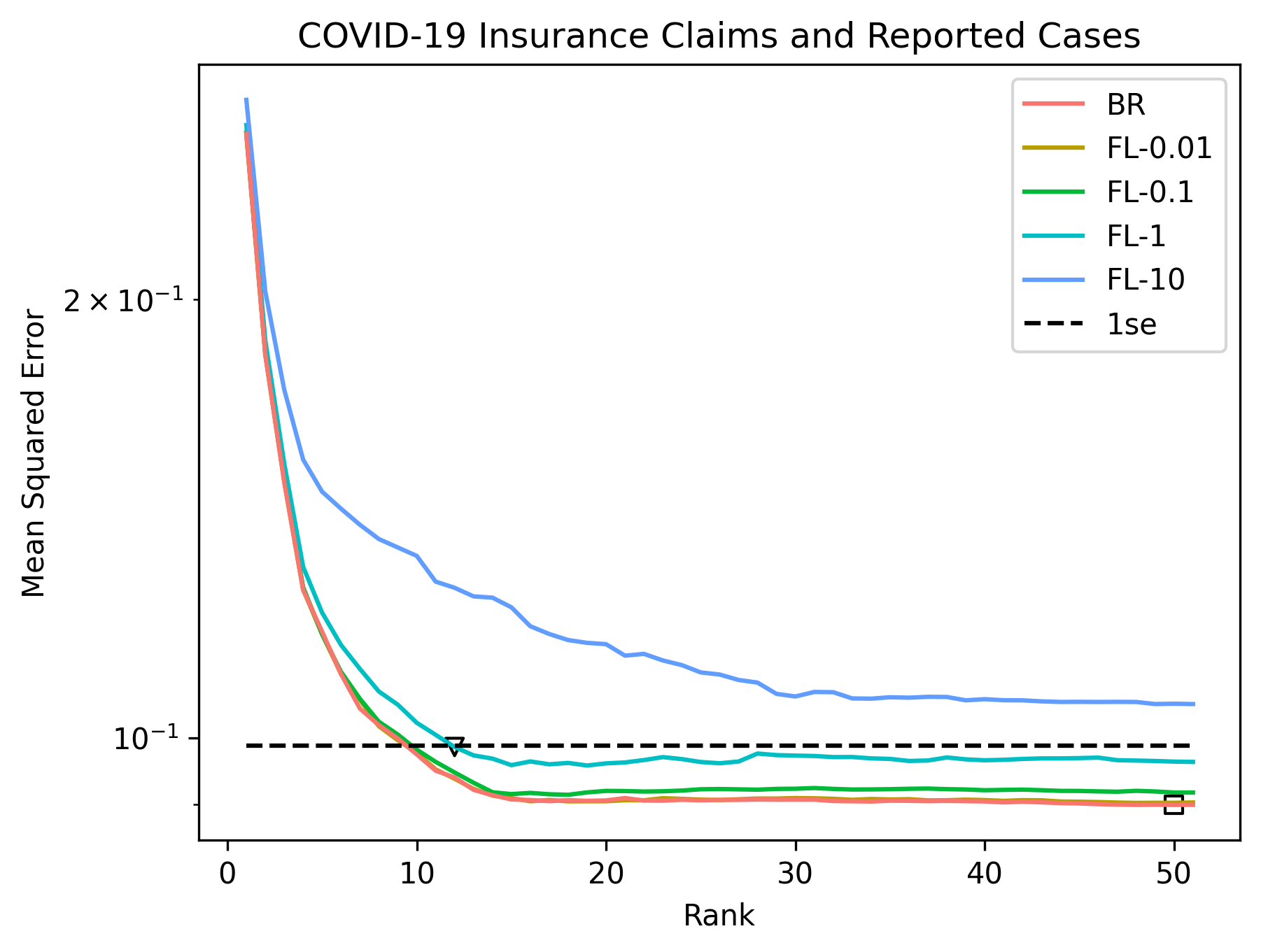}
    \caption{\small Cross validation error is optimized at $K=50$ using the Bounded Rank Model (BR), i.e. $\lambda=0$, indicated by the square. However, when applying the one standard error rule, we select the Fused Lasso Model (FL) with $K=12$ and $\lambda=1$, indicated by the triangle. This results in a great reduction in parameters with a small decrease in cross validation accuracy.}
    \label{fig:cv_covid}
\end{figure}

Unlike in the simulation experiments, we do not see a clear overfitting curve. As shown in Fig~\ref{fig:cv_covid}, the cross validation error decreases as $K$ increases and $\lambda$ decreases (as the model's complexity increases) and then flattens. The model with the best cross validation error has $K = 50$, where the rank of the difference matrix is $51$. Clearly, we do not want to use this model, since we do not believe that the heterogeneity present in the claims signal has rank $50$ out of a possible $51$. This is where the ``one standard error rule'' is useful. It selects the Fused Lasso Model with rank $K = 12$ and $\lambda = 1$. Although this model still has a higher rank than we may have thought appropriate, it is much simpler than the model which minimizes cross validation error.

\begin{figure}
    \centering
    \includegraphics[width=0.7\textwidth]{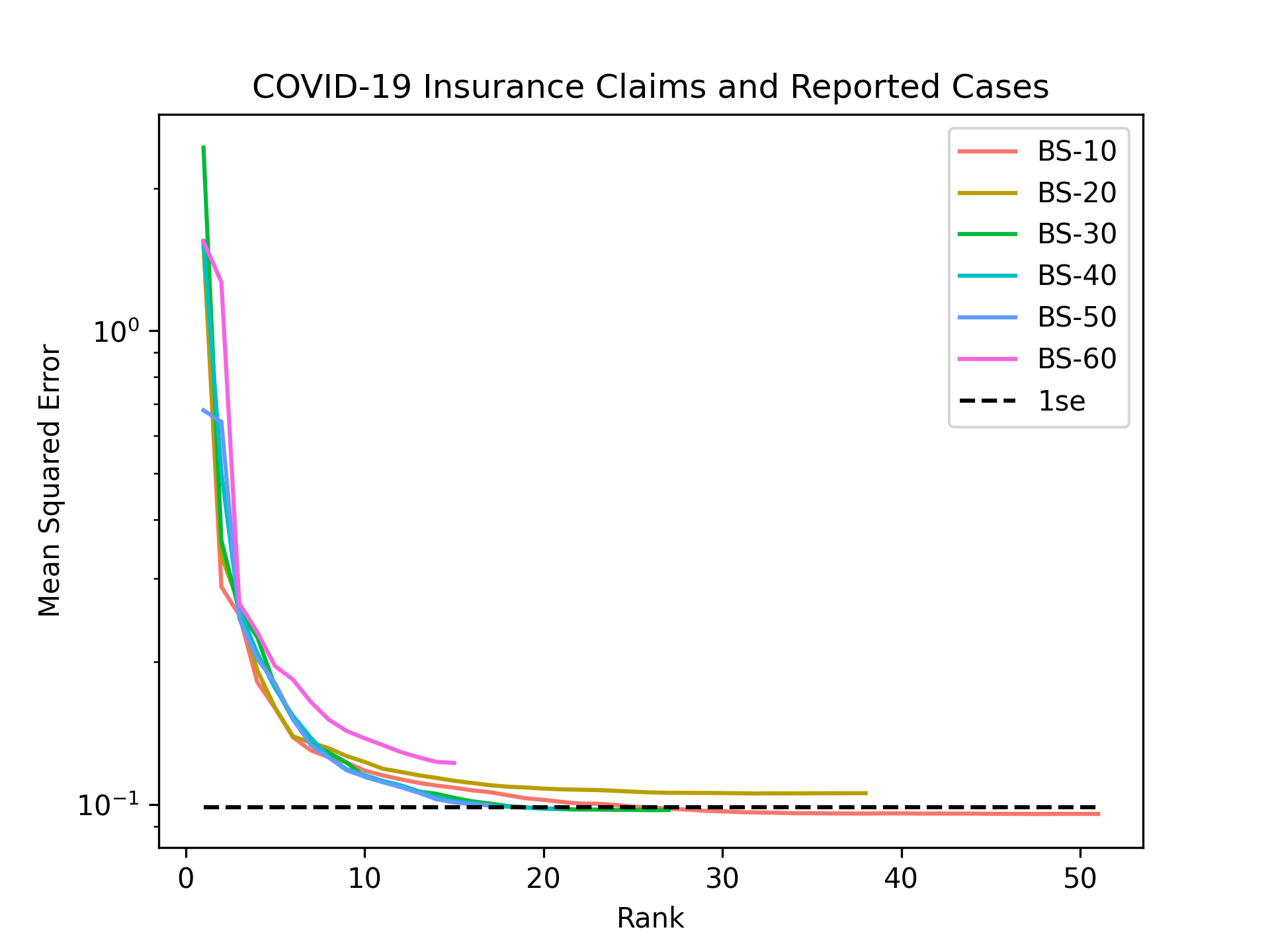}
    \caption{\small We plot the performance of the Basis Spline Model (BS) for different knot intervals. These models have a lower accuracy than the other models, but for some hyperparameters, the Basis Spline Model comes within one standard error of the best Bounded Rank or Fused Lasso Model. Note that the higher the knot interval, the lower the rank of the spline transformation matrix $C$. For knot intervals more than 10 days, the maximum rank of the model is less than $\min(N,T)=51$.}
    \label{fig:cv_covid_bs}
\end{figure}

\begin{figure}
    \centering
    \includegraphics[width=0.7\textwidth]{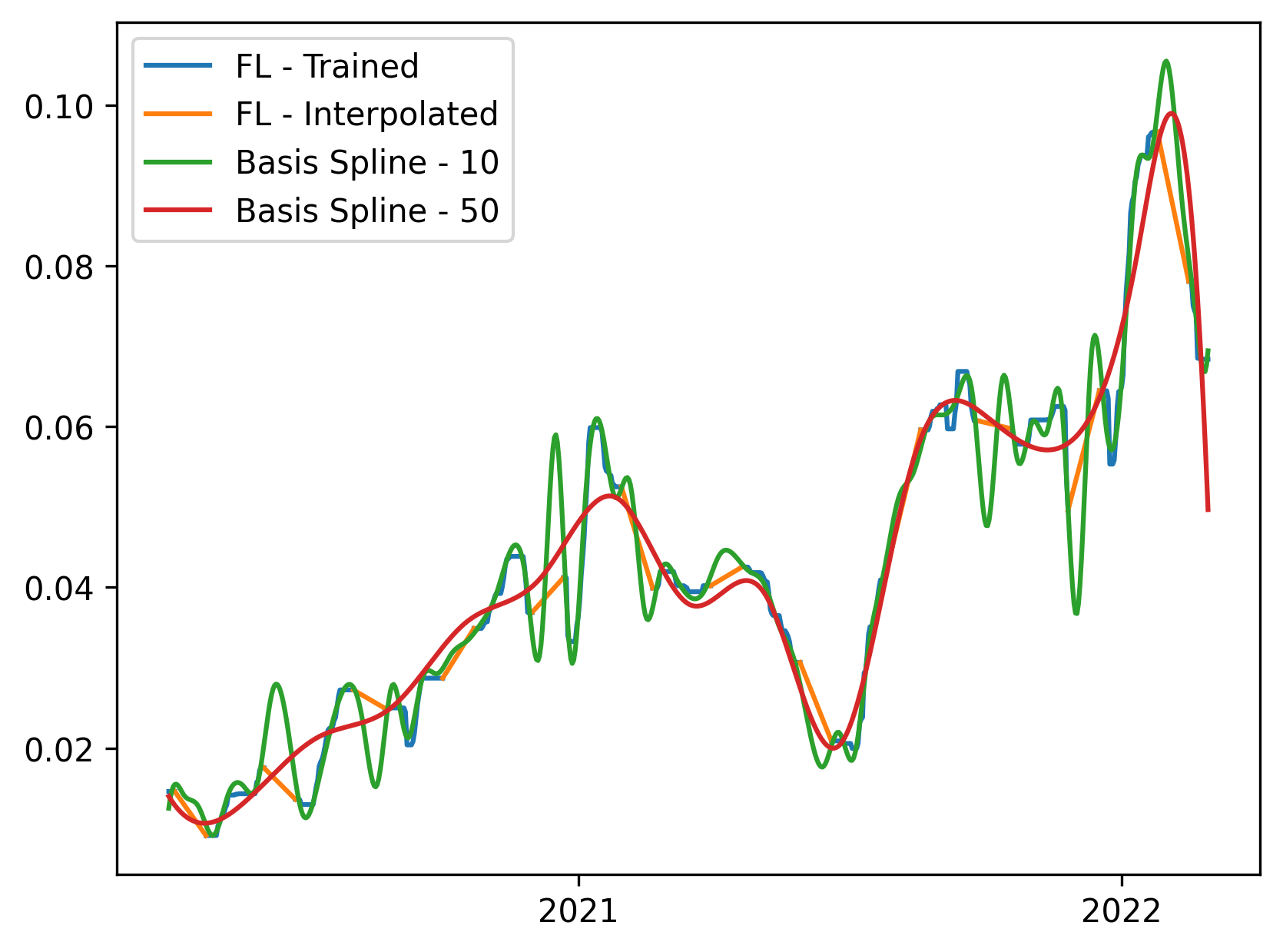}
    \caption{\small We plot the first component of the temporal correction matrix for the Fused Lasso (FL) and Basis Spline models. The orange segments correspond to held-out data that needs to be interpolated for the Fused Lasso Model. The Basis Spline Model with a knot interval of 10 tracks very closely to the Fused Lasso Model in training data but diverges wildly in held-out data. The Basis Spline Model with a knot interval of 50 is smooth but does not have the flexibility to closely match the Fused Lasso Model.}
    \label{fig:bs_oos}
\end{figure}

The Basis Spline Models perform poorly on this dataset, as shown in Fig~\ref{fig:cv_covid_bs}. For very small knot intervals, some models are candidates for selection under the ``one standard error rule'', but their degrees of freedom are larger than those of the Fused Lasso Models. We examine the behavior of the Basis Spline Models in Fig~\ref{fig:bs_oos}, where we see that there is overfitting if the knot interval is too short (many parameters) and underfitting if the knot interval is too long (fewer parameters). After performing linear interpolation, the overfitting model ends up with reasonable accuracy. However, the basis splines themselves do not accurately represent the temporal corrections. We conclude that the assumption of cubic splines is too rigid in this case. The splines simply cannot fit well to the data, likely due to abrupt changepoints that the Fused Lasso Models are able to handle better.


\begin{figure}
\centering
  \includegraphics[width=0.9\textwidth]{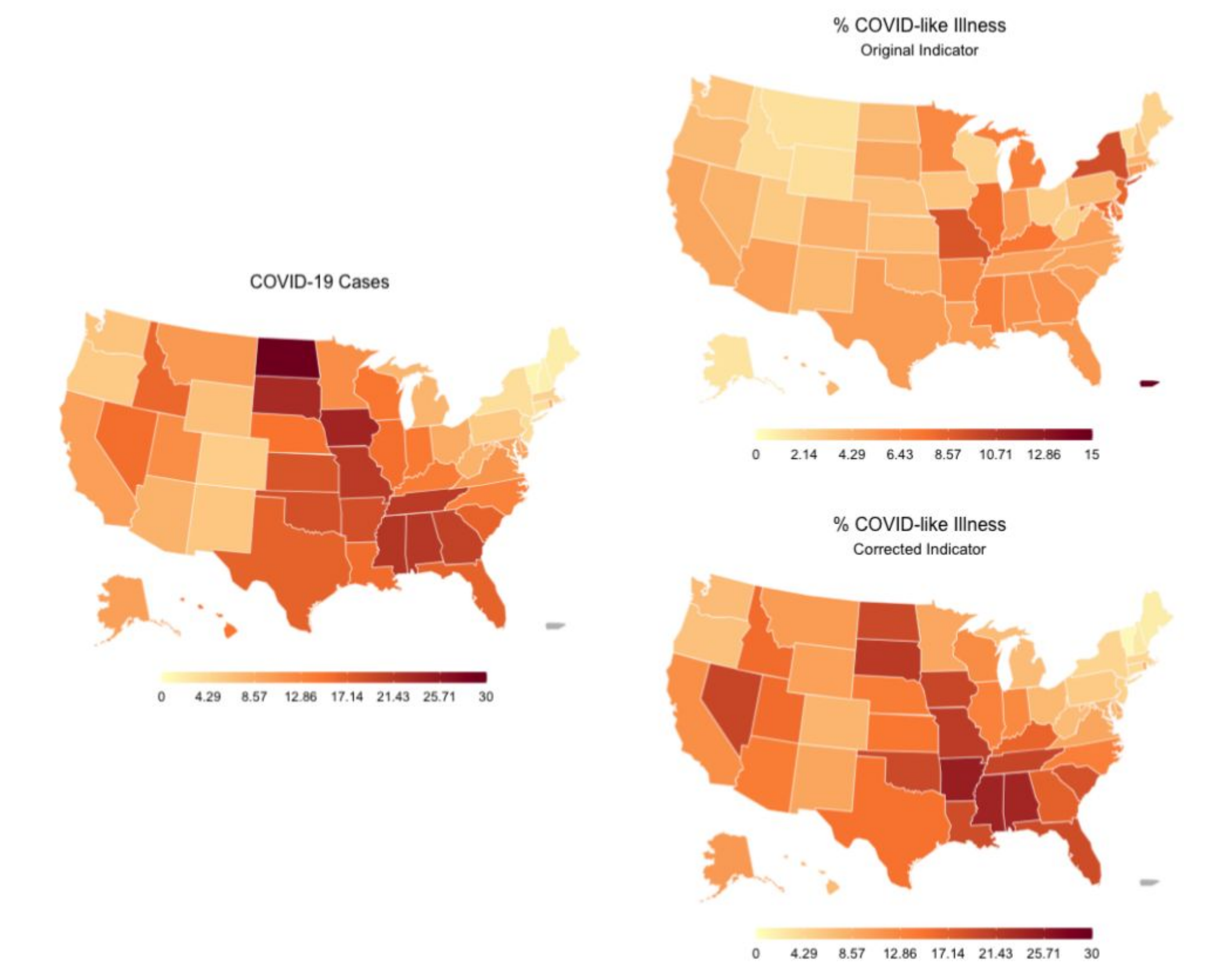}
    \caption{\small Applying a rank-2 correction improves similarity between reported COVID-19 cases and the insurance claims signal. On the left, the average daily confirmed COVID-19 cases between August 15 and September 15, 2020 are displayed in a choropleth map. On the right, we display the value of the insurance claims signal for the same time period before (top) and after (bottom) applying a rank-2 heterogeneity correction using the Bounded Rank Model. The pre-correction \%CLI map is not similar to the cases map, but the post-correction \%CLI map is.}
    \label{fig:choro_map}
\end{figure}

In Fig~\ref{fig:choro_map}, we illustrate the benefit of applying heterogeneity corrections. The raw insurance claims signal is quite different than the reported case signal in late summer in 2020. The state with the highest claims signal is New York, even though New York has one of the lowest rates of confirmed cases. After applying heterogeneity correction using cases as a guide, the insurance claims signal looks more similar to the reported case signal, improving the comparability of the insurance claims signal across states.

\subsection{Evaluating Preprocessing Assumptions}
As mentioned above, we applied a log transform to the data, assuming that the heterogeneity effects are multiplicative rather than additive. We can test that assumption by comparing the following three models.

\begin{enumerate}
    \item Bounded Rank Model with rank $k=1$ (BR-1): $$\min_{a,b} \sum_{i=1}^N \sum_{t=1}^T (\log X_{it} + a_i \cdot b_t - \log Y_{it})^2$$
    \item Additive Model in log space (AL): $$\min_{a,b} \sum_{i=1}^N \sum_{t=1}^T (\log X_{it} + a_i + b_t - \log Y_{it})^2$$
    \item Additive Model in count space (AC): $$\min_{a,b} \sum_{i=1}^N \sum_{t=1}^T \left(\frac{X_{it} + a_i + b_t}{Y_{it}} - 1\right)^2$$
\end{enumerate}

All of these models have $N+T$ parameters and a total of $N+T-1$ degrees of freedom, with a single parameter for each location and a single parameter for each day, with no regularization. In the first two models, the heterogeneity is assumed to be additive in the log space, or multiplicative in the count space. In the AC model, the heterogeneity is assumed to be additive in the count space. In the BR-1 model, the heterogeneity parameters are multiplied together, whereas in the AL model, they are added. Note that we minimize the relative error for the AC model so that all three models have the same objective.

We display the mean squared error between $\log \tilde{X}$ and $\log Y$ for each of the three models below, with the standard error in parentheses. The models BR-1 and AL, which assume that heterogeneity is multiplicative, perform much better than AC, which assumes that heterogeneity is additive. This supports our initial assumption that the effects of heterogeneity in this particular signal pair are multiplicative.

The AL model performs slightly better than the BR-1 model, which weakly suggests that the spatial and temporal parameters should be added instead of multiplied together. However, the AL model cannot be generalized to higher rank corrections. Therefore, we cannot use this model in practice, as we believe that the effects of heterogeneity are too complex to be modeled solely by a single parameter for each location and for each time point.

\begin{center}
\renewcommand*\arraystretch{1.3}
\begin{tabular}{|c|c|c|c|} \hline
     Model & BR-1 & AL & AC \\ \hline 
     MSE & 0.2205 (0.00220) & \textbf{0.2138 (0.00235)} & 0.7611 (0.00919) \\ \hline 
\end{tabular}
\end{center}

\subsection{Google Trends and CTIS Survey}
Google has made public an aggregated and anonymized dataset of Google search queries related to COVID-19 \cite{google2021}. An indicator derived from this dataset roughly measures the relative prevalence of a specific set of search queries in a given location and time. Ideally, this indicator could inform us approximately how many people are symptomatic at a given time at a very minimal cost. However, search query behavior is affected by many other factors other than whether a person is symptomatic. People may be more likely to search for COVID-19 related terms if someone famous was reported as infected, or if public health measures were enacted or lifted. These can create both spatial and temporal heterogeneity in the indicator.

We used the COVID-19 Trends and Impact Survey (CTIS) as a guide signal. Specifically, our guide is the estimated percentage of people with COVID-like illness from the survey. We used the hyperparameter selection scheme as above.

\begin{figure}
    \centering
    \includegraphics[width=0.7\textwidth]{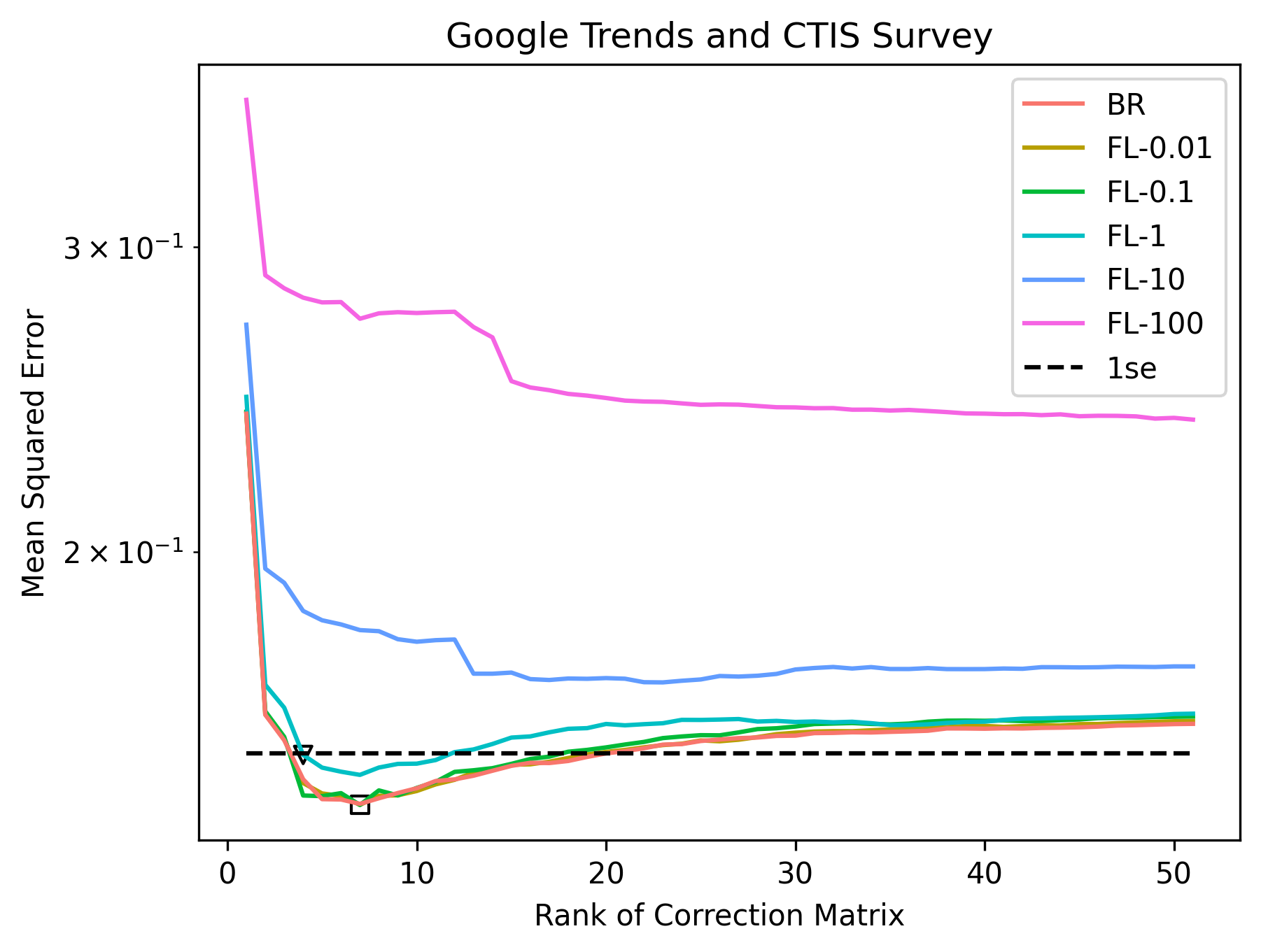}
    \caption{\small When correcting the Google Trends signal, cross validation error is optimized at $K=7$ using the Fused Lasso Model (FL) with $\lambda=0.1$, indicated by the square. However, when applying the one standard error rule, we select $K=4$ and $\lambda=1$, indicated by the triangle.}
    \label{fig:cv_ctis}
\end{figure}

The results here, shown in Fig~\ref{fig:cv_ctis}, look more similar to the results in the simulated dataset. The model that performs best in cross validation has a rank of $7$, and when applying the one standard error rule, the optimal model is a Fused Lasso Model with rank $K=4$ and $\lambda = 1$. With increasing $K$, the cross validation errors increase, indicating that some overfitting can occur. Here as well, the Basis Spline Models perform poorly (not pictured), entrenching a pattern seen in the insurance claims experiment as well.

\begin{figure}
    \centering
    \includegraphics[width=0.7\textwidth]{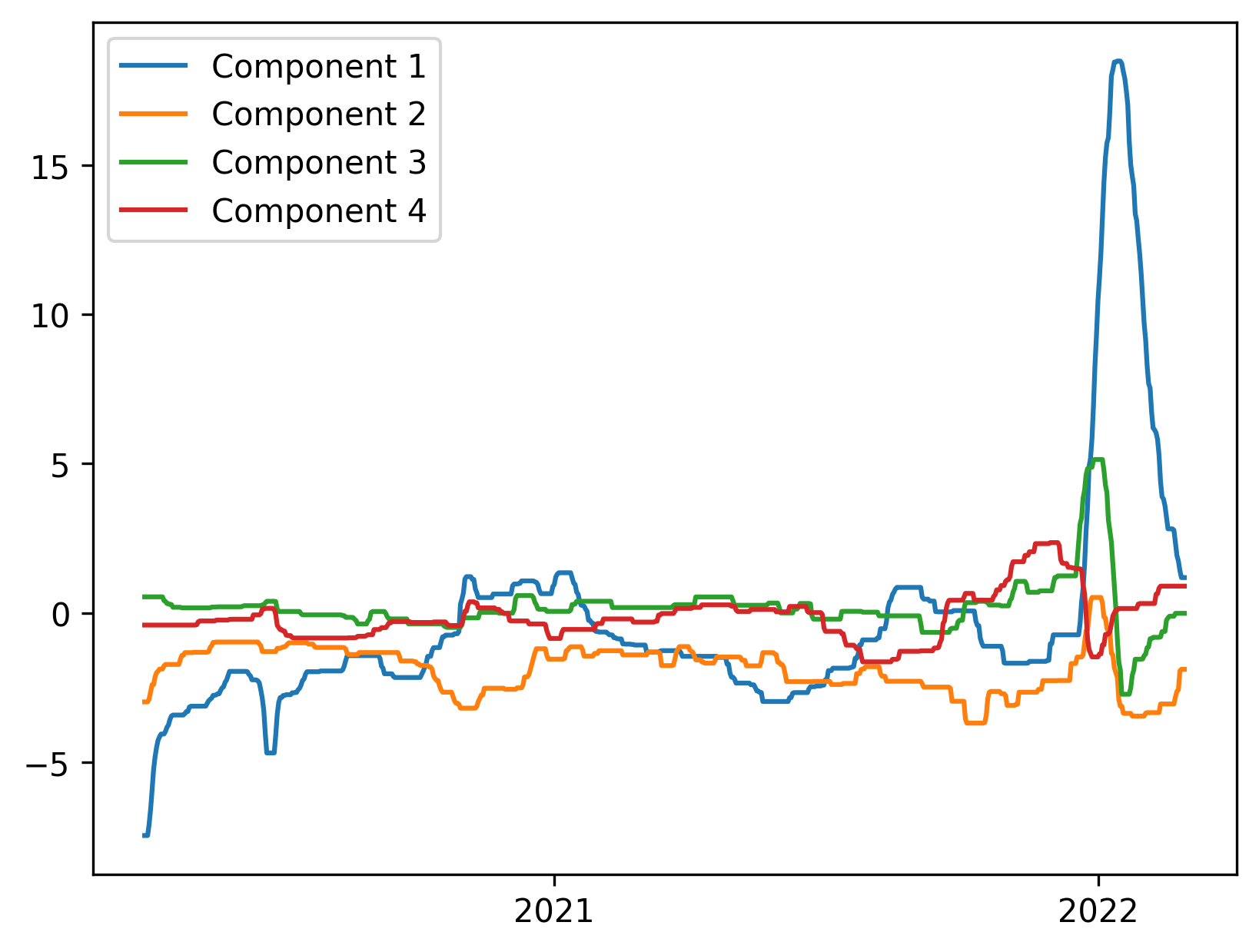}
    \caption{We plot the temporal components of the heterogeneity correction between the CTIS and Google Trends signal, using the Fused Lasso model with $K=4$ and $\lambda=1$. The most prominent corrections occur around January 2022, corresponding with the Omicron wave.}
    \label{fig:fl_components}
\end{figure}

We examine the temporal components of the optimal model in Fig~\ref{fig:fl_components}. As expected, the components are piecewise constant across time. The first (most important) component is mostly negative in the beginning of the pandemic and spikes during the Omicron wave. By using the CTIS survey as a guide, we correct the Google Trends signal downwards in the beginning of the pandemic and upwards during the Omicron wave.

One possible explanation for this heterogeneity is the decline in public attention and anxiety regarding the COVID-19 pandemic. In the beginning and middle of 2020, many asymptomatic people entered COVID-related searches into Google, resulting in a positively biased signal. Throughout most of 2021, minimal corrections are made and the two signals are at their strongest agreement. During the Omicron wave around the beginning of 2022, our method applies a strong positive correction to the Google Trends signal. According to the CTIS signal, COVID-19 cases are highest at this point, but the Google Trends signal does not increase to the same extent, so a further positive correction is needed. One possible explanation is that fewer symptomatic individuals were appearing in the Google signal, potentially because they were more confident that they indeed had a COVID-19 infection, or because they were less anxious. Another explanation could be that fewer non-symptomatic individuals were appearing in the Google signal, potentially because they were less interested in the pandemic. Whatever the exact reason, the corrections show that the Google signal suffers from temporal heterogeneity, which can be corrected by using the CTIS survey as a guide signal.

\section{Discussion}
As explained above, we define heterogeneity as the presence of location-dependent or time-dependent bias between an indicator and its unobserved ground truth. Indicators are useful sources of information for modeling, mapping, and forecasting epidemics, but conclusions derived from the indicators in the presence of heterogeneity may be suspect. The problem of heterogeneity is poorly suited to translate into an optimization problem in the absence of any ground truth data. Therefore, we use another signal as a guide, and present a method that can use the guide strongly or weakly.

Our method appears to be useful on several pairs of COVID-19 indicators. As Fig~\ref{fig:choro_map} shows, the raw COVID-19 insurance claims signal gives a very different picture than reported cases. If we were to use the insurance claims signal to understand the current COVID-19 burden across the United States, we could be very misinformed.

The flexibility of our approach is both its main strength and main weakness. On the one hand, the models discussed in this paper can be used for any generic signal and corresponding guide. The user can choose the appropriate parameters based on domain knowledge, exploratory data analysis (e.g. an elbow plot), or the cross validation scheme described above. Because heterogeneity is not straightforward to quantify, we require flexibility to cover a variety of use cases.

However, this flexibility requires a method to select hyperparameters. In simulations, cross validation yields a reasonable choice of hyperparameters. However, in a real-world setting, the hyperparameters selected by cross validation lead to a model that seems to overadjust. Cross validation might lead to model overadjustment because there are dependencies between the left-out data and the training data. In this case, just as we would expect to overfit in a normal prediction setup, we would expect to overadjust to the guide signal. Additionally, the error metric also encourages overadjustment, since we minimize the squared error between the corrected signal and the guide.

Another significant limitation of this approach is that the guide signal needs to be more reliable than the indicator we are trying to correct. Using Fig~\ref{fig:choro_map} as an example again, we see that $Y$ is low in New York but $X$ is high, and that after applying our heterogeneity correction, $\tilde{X}$ is low. This is only an improvement if $Y$ is correct, that true COVID-19 activity in New York is actually low. In this case, we have domain knowledge to suggest that reported cases suffer from spatial heterogeneity less than insurance claims. However, were we to treat cases as $X$ and insurance claims as $Y$, then our ``corrected'' case signal would be incorrectly high in New York.

An important extension to this approach would be modifying the hyperparameter selection scheme. A better scheme would not default to overadjustment so strongly and would not use an error metric that is optimized when fit exactly to the guide signal. Another extension would be the use of multiple guide signals $Y_1, \dots, Y_m$. A simple start would be to set $Y = \alpha_1 Y_1 + \dots  + \alpha_m Y_m$ and then apply the heterogeneity correction using $Y$ as the guide signal. Intuitively, if the sources of heterogeneity in the various guides are uncorrelated, then they will tend to cancel out as a result of this averaging, resulting in a spatially and temporally more homogeneous guide. Alternatively, we could view $X, Y_1, Y_2, \ldots, Y_m$ as $m+1$ different signals, and use them with the models discussed above to jointly estimate the underlying latent quantity to which they are all related. Using multiple guide signals will likely also reduce the overadjustment problem, and a more creative approach to incorporating multiple signals might avoid using the error with the guide signal as a performance metric for hyperparameter selection.

Our current setup fits the adjustment matrix in a batch setting, but a future direction would be to modify the algorithm in an online setting. Indicators are commonly used in real-time, so an online algorithm which makes adjustments as new data arrives may be more appropriate for many use cases.

Of the three models we propose, the Fused Lasso Model performs best in both simulated and real-world experiments. However, it is quite expensive computationally, whereas the other two models can be solved rapidly using SVD-based approaches. Given that the Bounded Rank Model usually performs well, it may be preferable to simply use the Bounded Rank Model in some applications. The Basis Spline Model is slightly more sophisticated without a meaningful increase in computation time. However, the assumptions that lie behind the Basis Spline Model seem to be too strong, specifically when there are abrupt changepoints in temporal heterogeneity.


\nolinenumbers
\bibliography{bib.bib}

\end{document}